%% file: main.tex
\documentclass[11pt,a4paper]{article}
\usepackage[hyperref]{emnlp2018}

\usepackage{times}
\usepackage{latexsym}
\usepackage{makecell}
\usepackage{graphicx}
\usepackage{subfigure}
\usepackage{algorithm}
\usepackage{amssymb}
\usepackage{amsmath}
\usepackage[noend]{algpseudocode}
\usepackage{multirow}
\usepackage{array}
\usepackage{tabularx}
\usepackage{booktabs}

\usepackage{url}

\input{defs}

\newcommand{\namecite}[1]{\newcite{#1}}

\usepackage{amsmath}
\usepackage{graphicx}
\usepackage{fancyhdr}
\usepackage{times}
\usepackage{epsfig}
\usepackage{graphicx}
\usepackage{amsmath}
\usepackage{amssymb}
\usepackage{multirow}
\usepackage{caption}
\usepackage{comment}
\begin{document}
\aclfinalcopy

\title{Ensemble Sequence Level Training for Multimodal MT: \\OSU-Baidu WMT18 Multimodal Machine Translation System Report}

\author{
  Renjie Zheng  \thanks{\quad Equal contribution} \hspace{0.5mm}$^1$
  \qquad
  Yilin Yang $^{* 1}$
  \qquad
  Mingbo Ma \thanks{\quad Contributions made while at Baidu Research} \hspace{0.05mm} $^{1,2}$
  \qquad
  Liang Huang  $^{\dag}$ $^{2,1}$
\\ 
$^{1}$School of EECS, Oregon State University, Corvallis, OR \\
$^{2}$Baidu Research, Sunnyvale, CA\\
  {\small \tt zheng@renj.me \quad \{yilinyang721, cosmmb, liang.huang.sh\}@gmail.com} \\
  }
\date{}
\maketitle

\begin{abstract}

This paper describes multimodal machine translation systems
developed jointly by Oregon State University and Baidu Research
for WMT 2018 Shared Task on multimodal translation.
In this paper, we introduce a simple approach
to incorporate image information by feeding image features to the decoder side.
We also explore different sequence level training methods including scheduled sampling
and reinforcement learning which lead to substantial improvements.
Our systems ensemble several models using different architectures
and training methods and achieve the best performance for three subtasks:
En-De and En-Cs in task 1 and (En+De+Fr)-Cs task 1B.

\end{abstract}

\input{intro}

\input{method}

\input{exps}

\section{Conclusions}

We describe our systems submitted to the shared WMT 2018 multimodal translation tasks.
We use sequence training methods which lead to substantial improvements over strong baselines.
Our ensembled models achieve the best performance in BLEU score for three subtasks: En-De, En-Cs of task 1 and (En+De+Fr)-Cs task 1B.

\section*{Acknowledgments}

This work was supported in part by DARPA grant N66001-17-2-4030, and NSF grants IIS-1817231 and IIS-1656051.
We thank the anonymous reviewers for suggestions and Juneki Hong for proofreading.

\bibliography{emnlp2018}
\bibliographystyle{acl_natbib_nourl}

\end{document}

%% file: defs.tex
\newcommand{\notes}[1]{}

\newcommand{\ith}[1]{\ensuremath{i^{{th}}}}

\newcount\permx
\newcount\permy
\def\permdot#1#2{
\permx=#1 \advance\permx by-1
\permy=#2 \advance\permy by-1
\psframe[fillcolor=black, fillstyle=solid]
(\permx,\permy)(#1, #2)
}

\newcommand{\boxnum}[1]{{\setlength{\fboxsep}{1pt}\raisebox{1pt}{\hspace{1pt}\fbox{\tiny #1}\hspace{1pt}}}}
\newcommand{\ind}[1]{\ensuremath{_{\kern-0.5pt\boxnum{#1}}}}

\newcommand{\smallnt}[1]{\ensuremath{_{\mbox{\tiny PP}}}\xspace}

\newcommand{\pseudocode}{Algorithm}
\floatname{algorithm}{\pseudocode}

\iffalse

\else

\fi



%% file: intro.tex
\section{Introduction}

In recent years, neural text generation has attracted much attention due to its impressive generation accuracy and wide applicability. 
In addition to demonstrating compelling results for machine translation
\cite{Sutskever:2014, Bahdanau:14}, by simple adaptation,
similar models have also proven to be successful for
summarization \cite{Rush:15,Nallapati:16},
image or video captioning \cite{Venu:15,Kelvin:15}
and multimodal machine translation \cite{elliott2017findings, caglayan2017lium,calixto2017incorporating,ma2017osu}, 
which aims to translate the caption from one language to another
with the help of the corresponding image.

However, the conventional neural text generation models
suffer from two major drawbacks.
First, they are typically trained by predicting
the next word given the previous ground-truth word.
But at test time, the models recurrently feed their own predictions into it.
This ``exposure bias'' \cite{ranzato2015sequence} leads to error accumulation during generation at test time.
Second, the models are optimized by
maximizing the probability of the next ground-truth words
which is different from
 the desired non-differentiable evaluation metrics, e.g.~BLEU.

Several approaches have been proposed to tackle the previous problems.
\namecite{bengio2015scheduled} propose scheduled sampling
 to alleviate ``exposure bias'' by
feeding back the model's own predictions
with a slowly increasing probability during training.
Furthermore, reinforcement learning \cite{sutton1998reinforcement} is
proven to be helpful to directly optimize the evaluation metrics 
in neural text generation models training.
\namecite{ranzato2015sequence} successfully use the REINFORCE algorithm to directly optimize the evaluation metric over multiple text generation tasks.
\namecite{rennie2017self, liu2017improved} achieve state-of-the-art
on image captioning using REINFORCE with baseline to reduce training variance.


Moreover,
many existing works show that neural text generation models
can benefit from model ensembling 
by simply averaging the outputs of different models
\cite{elliott2017findings,rennie2017self}.
\namecite{garmash2016ensemble} claim that
it is essential to introduce diverse models into the ensemble.
To this end, we ensemble models
with various architectures and training methods.

This paper describes our participation in the WMT 2018 multimodal tasks. Our submitted systems include a series of models which 
only consider text information,
as well as multimodal models which also include image information
to initialize the decoders.
We train these models using scheduled sampling and reinforcement learning.
The final outputs are decoded by ensembling those models.
To the best of our knowledge, this is the first multimodal
machine translation system that achieves the state-of-the-art
using sequence level learning methods.

%% file: method.tex

\section{Methods}

\begin{figure}
\includegraphics[width=0.5\textwidth]{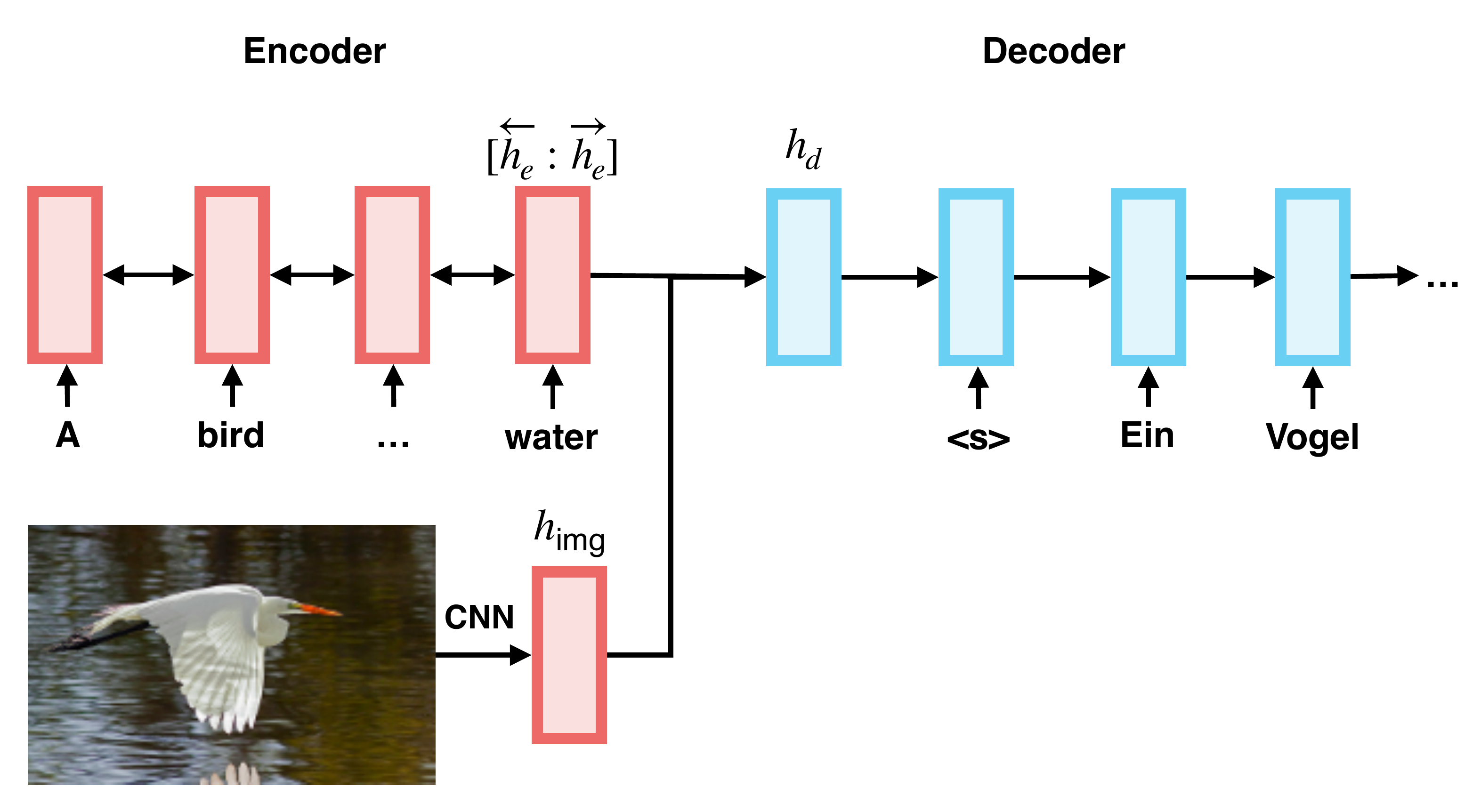}
\caption{Multimodal Machine Translation Model}
\label{fig:model}
\end{figure}

Our model is based on the sequence-to-sequence RNN architecture
with attention \cite{Bahdanau:14}.
We incorporate image features to initialize
the decoder's hidden state as shown in Figure \ref{fig:model}.
Originally, this hidden state is initialized using
the concatenation of last encoder's
forward and backward hidden states,
$\overrightarrow{h_e}$ and $\overleftarrow{h_e}$ resp.
We propose to use the sum of encoder's output and
image features $h_{\text{img}}$ to initialize the decoder.
Formally, we have the final initialization state $h_d$ as:
\begin{equation}
h_d = \tanh(W_{e}[\overrightarrow{h_e}; \overleftarrow{h_e}] + W_{\text{img}} h_{\text{img}} + b) .
\end{equation}
where $W_{e}$ and $W_{\text{img}}$ project the encoder and image 
feature vector into the decoder hidden state dimensionality and
$b$ is the bias parameter.
This approach has been previously explored by \namecite{calixto2017incorporating}.

As discussed previously, translation systems are traditionally
trained using cross entropy loss.
To overcome the discrepancy between training and inference distributions,
we train our models using scheduled sampling \cite{bengio2015scheduled} which 
mixes the ground truth with model predictions,
further adopting the REINFORCE algorithm with baseline to
directly optimize translation metrics.

\subsection{Scheduled Sampling}


When predicting a token $\hat{y}_t$, scheduled sampling uses
the previous model prediction
$\hat{y}_{t-1}$ with probability $\epsilon$ or the previous ground truth
prediction $y_{t-1}$ with probability $1 - \epsilon$. 
The model prediction $\hat{y}_{t-1}$ is obtained by sampling a token
according to the probability distribution by $P(y_{t-1}|h_{t-1})$.
At the beginning of training, the sampled token can be very random.
Thus, the probability $\epsilon$ is set very low 
initially and increased over time.

One major limitation of scheduled sampling is that at each time step,
the target sequences can be incorrect since
they are randomly selected from the ground truth data or model predictions,
regardless of how input was chosen \cite{ranzato2015sequence}.
Thus, we use reinforcement learning techniques to further optimize
models on translation metrics directly.

\subsection{Reinforcement Learning}

Following \namecite{ranzato2015sequence} and \namecite{rennie2017self},
we use REINFORCE with baseline to directly optimize
the evaluation metric.


According to the reinforcement learning literature \cite{sutton1998reinforcement},
the neural network, $\theta$, defines a policy $p_{\theta}$,
that results in an ``action'' that is the prediction of next word.
After generating the end-of-sequence term (EOS),
the model will get a reward $r$, which can be the evaluation metric,
e.g. BLEU score, between the golden and generated sequence.
The goal of training is to minimize the negative expected reward.
\begin{equation}
L(\theta)= -  \mathbb{E}_{w^s \sim p_{\theta}} \left[ r(w^s)\right].
\end{equation}
where sentence $w^s = (w_1^s, ..., w_T^s)$.

In order to compute the gradient $\nabla_{\theta}L(\theta)$, we use the REINFORCE algorithm, which is based on the observation that the expected gradient of a non-differentiable reward function can be computed as follows:
\begin{equation}
\nabla_{\theta}L(\theta)= - \mathbb{E}_{w^s \sim p_{\theta}} \left[ r(w^s) \nabla_{\theta} \log p_{\theta}(w^s) \right].
\end{equation}

The policy gradient can be generalized to compute the reward associated with an action value \emph{relative} to a reference reward or \emph{baseline} $b$:
\begin{equation}
\nabla_{\theta}L(\theta)= - \mathbb{E}_{w^s \sim p_{\theta}} \left[ (r(w^s)- b ) \nabla_{\theta} \log p_{\theta}(w^s)\right].
\end{equation}

The baseline does not change the expected gradient, but importantly,
it can reduce the variance of the gradient estimate. 
We use the baseline introduced in \namecite{rennie2017self} which
is obtained by the current model with greedy decoding at test time.
\begin{equation}
b=r(\hat{w^s})
\end{equation}
where $\hat{w^s}$ is generated by greedy decoding.

For each training case, we approximate the expected gradient with a single sample $w^s\sim p_{\theta}$:
\begin{equation}
\nabla_{\theta}L(\theta)\approx- (r(w^s)-b) \nabla_{\theta} \log p_{\theta}(w^s).
\label{eq:zerobiasGradientSample}
\end{equation}

\subsection{Ensembling}

In our experiments with relatively small training dataset,
the translation qualities of models with different initializations
can vary notably.
To make the performance much more stable and improve the translation quality,
we ensemble different models during decoding to achieve better translation.

To ensemble, we take the average of all model outputs:
\begin{equation}
\hat{y_t} = \sum_{i = 1}^N {\hat{y^i_t} \over N}
\end{equation}
where $\hat{y^i_t}$ denotes the output distribution of $i$th model
at position $t$.
Similar to \namecite{zhou2017neural}, we can ensemble models
trained with different architectures and training algorithms.

%% file: exps.tex

\section{Experiments}


\subsection{Datasets}

\begin{table}[!]\centering
\begin{tabular}{|l|c|c|c|c|}\hline
   & Train & Dev. & Vocab. & Vocab. after BPE\\\hline
En & 2,900  & 1,014 & 10,212        & 7,633      \\\hline
De & 2,900  & 1,014 & 18,726        & 5,942      \\\hline
Fr & 2,900  & 1,014 & 11,223        & 6,457      \\\hline
Cs & 2,900  & 1,014 & 22,400        & 8,459      \\\hline
\end{tabular}
\caption{Statistics of Flickr30K Dataset}
\label{tab:dataset}
\end{table}

We perform experiments using Flickr30K \cite{elliott:2016}
which are provided by the WMT organization.
Task 1 (Multimodal Machine Translation) consists of 
translating an image with an English caption
into German, French and Czech.
Task 1b (Multisource Multimodal Machine Translation)
involves translating parallel English, German and French sentences
with accompanying image into Czech.

As shown in Table \ref{tab:dataset}, both tasks have
2900 training and 1014 validation examples.
For preprocessing, we convert all of the sentences to lower case,
normalize the punctuation, and tokenize.
We employ byte-pair encoding (BPE) \cite{sennrich2015neural}
on the whole training data including the four languages
 and reduce the source and target language vocabulary sizes to 20k
in total.

\subsection{Training details}

The image feature is extracted
using ResNet-101 \cite{kaiming:2016}
convolutional neural network trained on the ImageNet dataset.
Our implementation is adapted from Pytorch-based OpenNMT \cite{2017opennmt}.
We use two layered bi-LSTM \cite{Sutskever:2014} as the encoder and
share the vocabulary between the encoder and the decoder.
We adopt length reward \cite{huang+:2017} on En-Cs task to find
the optimal sentence length.
We use a batch size of 50, SGD optimization,
dropout rate as 0.1 and learning rate as 1.0.
Our word embeddings are randomly initialized of dimension 500.

To train the model with scheduled sampling,
we first set probability $\epsilon$ as 0,
and then gradually increase it 0.05 every 5 epochs until it's 0.25.
The reinforcement learning models are trained based on those
models pre-trained by scheduled sampling.

\subsection{Results for task 1}

\begin{table}[!]\centering
\begin{tabular}{|l|c|c|c|}\hline
             & En-De & En-Fr & En-Cs \\\hline
NMT          & 39.64 & 58.36 & 31.27 \\\hline
NMT+SS       & 40.19 & 58.67 & 31.38 \\\hline
NMT+SS+RL    & 40.60 & 58.80 & 31.73 \\\hline
MNMT         & 39.27 & 57.92 & 30.84 \\\hline
MNMT+SS      & 39.87 & 58.80 & 31.21 \\\hline
MNMT+SS+RL   & 40.39 & 58.78 & 31.36 \\\hline
NMT Ensemble & \bf{42.54} & 61.43 & \bf{33.15} \\\hline
MIX Ensemble & 42.45 & \bf{61.45} & 33.11 \\\hline
\end{tabular}
\caption{BLEU scores of different approaches on the validation set.
Details of the ensemble models are described in Table \ref{tab:ensemble}.}
\label{tab:dev_results}
\end{table}

\begin{table}[b]
\centering
\resizebox{0.5\textwidth}{!}{
\begin{tabular}{|l|l|c|c|c|c|} \hline
Task                   & System     & NMT+SS & NMT+SS+RL & MNMT+SS & MNMT+SS+RL \\ \hline
\multirow{2}{*}{En-De} & NMT & 7      & 6        & 0       & 0   \\ \cline{2-6}
                       & MIX & 7      & 6        & 5       & 4     \\ \hline
\multirow{2}{*}{En-Fr} & NMT & 9      & 5        & 0      & 0 \\ \cline{2-6}
                       & MIX & 9      & 0        & 3       & 0     \\ \hline
\multirow{2}{*}{En-Cs} & NMT & 7      & 6        & 0      & 0 \\ \cline{2-6}
                       & MIX & 7      & 6        & 5          & 4     \\ \hline
\end{tabular}}
\caption{Number of different models used for ensembling.}
\label{tab:ensemble}
\end{table}

To study the performance of different approaches,
we conduct an ablation study.
Table \ref{tab:dev_results} shows the BLEU scores
on validation set with different models and training methods.
Generally, models with scheduled sampling perform better than
baseline models, and reinforcement learning further improves the performance.
Ensemble models lead to substantial improvements over the best single model
by about +2 to +3 BLEU scores.
However, by including image information, MNMT performs better than NMT
only on the En-Fr task with scheduled sampling.

\begin{table}[!]
\resizebox{0.5\textwidth}{!}{
\begin{tabular}{|l|c|c|c|c|}\hline
             & Rank & BLEU & METEOR & TER \\\hline
OSU-BD-NMT   & 1    & \bf{32.3} & 50.9   & 49.9 \\\hline
OSU-BD-MIX   & 2    & 32.1 & 50.7   & \bf{49.6} \\\hline
LIUMCVC-MNMT-E& 3    & 31.4 & 51.4   & 52.1 \\\hline
UMONS-DeepGru & 4    & 31.1 & \bf{51.6}   & 53.4 \\\hline
LIUMCVC-NMT-E & 5    & 31.1 & 51.5   & 52.6 \\\hline
SHEF1-ENMT  & 6    & 30.9 & 50.7   & 52.4 \\\hline
Baseline     & -    & 27.6 & 47.4   & 55.2 \\\hline
\end{tabular}
}
\caption{En-De results on test set. 17 systems in total. (Only including constrained models).}
\label{tab:en_de_results}
\end{table}


\begin{table}[!]
\resizebox{0.5\textwidth}{!}{
\begin{tabular}{|l|c|c|c|c|}\hline
             & Rank & BLEU & METEOR & TER \\\hline
LIUMCVC-MNMT-E& 1    & \bf{39.5} & 59.9  & 41.7 \\\hline
UMONS & 2    & 39.2 & \bf{60}   & 41.8 \\\hline
LIUMCVC-NMT-E & 3    & 39.1 & 59.8   & 41.9 \\\hline
OSU-BD-NMT & 4    & 39.0 & 59.5   & \bf{41.2} \\\hline
SHEF-MLT & 5    & 38.9 & 59.8   & 41.5 \\\hline
OSU-BD-MIX & 9    & 38.6 & 59.3 & 41.5 \\\hline
Baseline     & -    & 28.6 & 52.2 & 58.8 \\\hline
\end{tabular}
}
\caption{En-Fr results on test set. 14 systems in total. (Only including constrained models).}
\label{tab:en_fr_results}
\end{table}



\begin{table}[!]
\resizebox{0.5\textwidth}{!}{
\begin{tabular}{|l|c|c|c|c|}\hline
             & Rank & BLEU & METEOR & TER \\\hline
OSU-BD-NMT   & 1    & \bf{30.2} & 29.5   & \bf{50.7} \\\hline
OSU-BD-MIX   & 2    & 30.1 & \bf{29.7}   & 51.2 \\\hline
SHEF1-ENMT  & 3    & 29.0 & 29.4   & 51.1 \\\hline
SHEF-LT     & 4    & 28.3 & 29.1   & 51.7 \\\hline
SHEF-MLT    & 5    & 28.2 & 29.1   & 51.7 \\\hline
SHEF1-MFS   & 6    & 27.8 & 29.2   & 52.4 \\\hline
Baseline     & -    & 26.5 & 27.7   & 54.4 \\\hline
\end{tabular}
}
\caption{En-Cs results on test set. 8 systems in total. (Only including constrained models).}
\label{tab:en_cs_results}
\end{table}

Table \ref{tab:en_de_results}, \ref{tab:en_fr_results} and \ref{tab:en_cs_results}
show the test set performance of our models on En-De, En-Fr and En-Cs subtasks with other top performance models.
We rank those models according to BLEU.
Our submitted systems rank first in BLEU and TER on En-De and En-Cs subtasks.

\subsection{Results for task 1B}

\begin{table}[!]\centering
\resizebox{0.5\textwidth}{!}{
\begin{tabular}{|l|c|c|c|c|}\hline
             & En-Cs & Fr-Cs & De-Cs & (En+Fr+De)-Cs \\\hline
NMT          & \bf{31.27} & 28.48 & 26.96 & 29.47 \\\hline
MNMT         & 30.84 & 27.02 & 25.99 & 29.23 \\\hline
\end{tabular}
}
\caption{BLEU scores on validation set for task 1B}
\label{tab:1b_results}
\vspace{-5.0mm}
\end{table}

\begin{table}[!]\centering
\resizebox{0.5\textwidth}{!}{
\begin{tabular}{|l|c|c|c|c|}\hline
             & Rank & BLEU & METEOR & TER \\\hline
OSU-BD-NMT   & 1    & \bf{26.4} & 28.0   & \bf{52.1} \\\hline
OSU-BD-MIX   & 1    & \bf{26.4} & \bf{28.2}   & 52.7 \\\hline
SHEF1-ARNN  & 3    & 25.2 & 27.5   & 53.9 \\\hline
SHEF-CON    & 4    & 24.7 & 27.6   & \bf{52.1} \\\hline
SHEF-MLTC   & 5    & 24.5 & 27.5   & 52.5 \\\hline
SHEF1-ARF   & 6    & 24.1 & 27.1   & 54.6 \\\hline
Baseline     & -    & 23.6 & 26.8   & 54.2 \\\hline
\end{tabular}
}
\caption{Task 1B multi-source translation results on test set. 6 systems in total.}
\label{tab:1b_test_results}
\end{table}

Table \ref{tab:1b_results} shows the results on validation set
without sequence training.
En-Cs, Fr-Cs, De-Cs are models trained from one language to another.
(En+Fr+De)-Cs models are trained using multiple source data.
Similar to the Shuffle method discussed in 
multi-reference training \cite{zheng2018multi}, 
we randomly shuffle the source data in all languages
and train using a traditional attention based-neural machine translation model in every epoch.
Since we do BPE on the whole training data, we can share the vocabulary
of different languages during training.
The results show that models trained using single English to Czech data
perform much better than the rest.

Table \ref{tab:1b_test_results} shows results on test set.
The submitted systems are the same as those used in En-Cs task of task 1.
Although we only consider the English source during training,
our proposed systems still rank first among all the submissions.

\begin{table}[tb]
\centering
\resizebox{0.5\textwidth}{!}{
\begin{tabular}{|l|l|c|c|c|c|c|c|c|c|} \hline
        \multirow{2}{*}{Task}&\multirow{2}{*}{System}         & \multicolumn{4}{c|}{Model Rank} &\multicolumn{4}{c|}{Team Rank} \\ \cline{3-10}
& & Num $^\dagger$ & BLEU & MET. & TER & Num $^\ddagger$ & BLEU  & MET. & TER \\ \hline
\multirow{2}{*}{En-De} & NMT & 11 & 1 & 4 & 2&\multirow{2}{*}{5}&\multirow{2}{*}{1}&\multirow{2}{*}{3}&\multirow{2}{*}{1}\\ \cline{2-6}
                       & MIX & 11 & 2 & 5 & 1& &      &       & \\ \hline
\multirow{2}{*}{En-Fr} & NMT & 11 & 4 & 9 & 1&\multirow{2}{*}{6}&\multirow{2}{*}{3}&\multirow{2}{*}{5}&\multirow{2}{*}{1} \\ \cline{2-6}
                       & MIX & 11 & 9 & 10  & 3& &    &       &  \\ \hline
\multirow{2}{*}{En-Cs} & NMT & 6 & 1  & 1 & 1 &\multirow{2}{*}{3}&\multirow{2}{*}{1}&\multirow{2}{*}{1}&\multirow{2}{*}{1}\\ \cline{2-6}
                       & MIX & 6 & 2  & 2 & 3 & &   &       &  \\ \hline
En-Cs&NMT&6&1& 2&1&\multirow{2}{*}{3}&\multirow{2}{*}{1}&\multirow{2}{*}{1}&\multirow{2}{*}{1}\\ \cline{2-6}
   (1B)                    & MIX &6&1& 1 & 5& &  &      & \\ \hline
\end{tabular}}
\caption{Rank of our models. $^\dagger$ represents the total number of models. $^\ddagger$ represents the total number of teams.}
\label{tab:ensemble}
\vspace{-5.0mm}
\end{table}